\documentclass[10pt,twocolumn,letterpaper]{article}

\usepackage{wacv}

\usepackage[numbers]{natbib}

\usepackage[utf8]{inputenc} 
\usepackage[T1]{fontenc}    
\usepackage{url}            
\usepackage{booktabs}       
\usepackage{amsfonts}       
\usepackage{nicefrac}       
\usepackage{microtype}      

\usepackage{comment}
\usepackage{epsfig}
\usepackage{epstopdf}
\usepackage{graphics}
\usepackage{subfigure}

\usepackage{times}
\usepackage[table]{xcolor}
\usepackage{soul}
\usepackage[utf8]{inputenc}
\usepackage{caption}

\usepackage{amsmath}

\usepackage{flushend}

\definecolor{lightgray}{gray}{0.85}
\DeclareMathOperator*{\argmin}{arg\,min}

\newif\ifcommenton
\commentontrue
\ifcommenton
\newcommand{\red}[1]{\textcolor{red}{#1} }
\else
\newcommand{\red}[1]{}
\fi

\wacvfinalcopy 

\def\wacvPaperID{411} 

\ifwacvfinal\fi
\begin{document}

\title{Uncertainty-aware Short-term Motion Prediction \\of Traffic Actors for Autonomous Driving}


\author{Nemanja Djuric, Vladan Radosavljevic, Henggang Cui, Thi Nguyen, \\Fang-Chieh Chou, Tsung-Han Lin, Nitin Singh, Jeff Schneider\\
Uber Advanced Technologies Group\\
{\tt\small \{ndjuric, vradosavljevic, hcui2, thi, fchou, hanklin, nitin.singh, jschneider\}@uber.com}
}

\maketitle
\ifwacvfinal\thispagestyle{empty}\fi

\begin{abstract}
We address one of the crucial aspects necessary for safe and efficient operations of autonomous vehicles, namely predicting future state of traffic actors in the autonomous vehicle's surroundings. We introduce a deep learning-based approach that takes into account a current world state and produces raster images of each actor's vicinity. The rasters are then used as inputs to deep convolutional models to infer future movement of actors while also accounting for and capturing inherent uncertainty of the prediction task. Extensive experiments on real-world data strongly suggest benefits of the proposed approach. Moreover, following completion of the offline tests the system was successfully tested onboard self-driving vehicles.

\end{abstract}

\vspace{-0.2cm}

\section{Introduction}

Driving a motor vehicle is a complex undertaking, requiring drivers to understand involved multi-actor scenes in real time and act upon rapidly changing environment within a fraction of a second ({\it actor} is a term referring to any vehicle, pedestrian, bicycle, or other potentially moving object). Unfortunately, humans are infamously ill-fitted for the task, as sadly corroborated by grim road statistics that often worsen year after year. Traffic accidents were the number four cause of death in the US in 2015, accounting for more than $5\%$ of the total \cite{national2017health}. In addition, despite large investments by governments and progress made in traffic safety technologies, in the US the year 2017 was still one of the deadliest years for motorists in the past decade \cite{nhtsa2017}. Moreover, human error is responsible for up to $94\%$ of crashes \cite{singh2015critical}, suggesting that removing the unreliable human factor could potentially save hundreds of thousands of lives and tens of billions of dollars in accident-related damages and medical expenses \cite{national2014economic}.

Latest breakthroughs in AI and high-performance computing, delivering powerful hardware at lower costs, unlocked the potential to reverse the negative safety trend on our public roads. In particular, together they gave rise to a development of the self-driving technology, where driving decisions are entrusted to a computer aboard a self-driving vehicle (SDV), equipped with a number of external sensors and capable of processing large amounts of information at speeds and throughputs far surpassing human capabilities. Once mature the technology is expected to drastically improve road safety and redefine the very way we organize transportation and our lives \cite{polzin2016implications}. To this end, the industry and governments are working closely to fulfill this potential and bring the SDVs to consumers, with companies such as Waymo, Uber, and Lyft investing significant resources into autonomous research, and states such as Texas, Pennsylvania, and California enacting necessary legal frameworks. Nevertheless, autonomous driving is still in initial development phases, with a number of challenges lying ahead of the researchers. 

\begin{figure*}[t!]
\centering
\includegraphics[keepaspectratio=1,width=0.77\columnwidth]{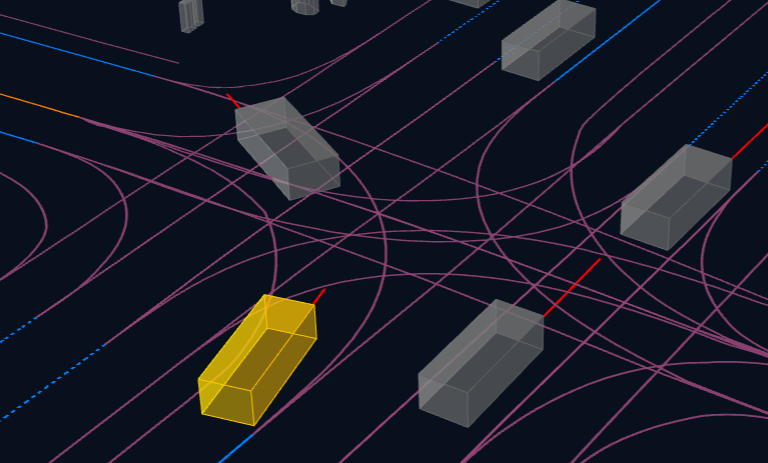}
\includegraphics[keepaspectratio=1,width=0.57\columnwidth,trim={0 1cm 0cm 0cm},clip]{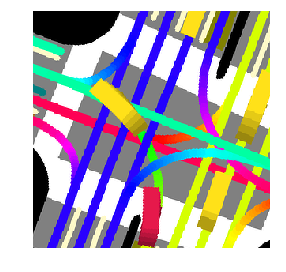}
\includegraphics[keepaspectratio=1,width=0.57\columnwidth,trim={0 1cm 0cm 0cm},clip]{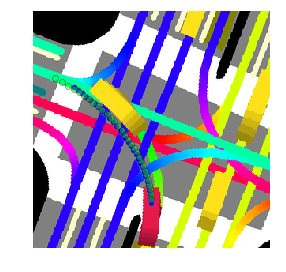}
\caption{Complex intersection scene handled by our model; (a) scene in a 3D viewer, with lane boundaries, surrounding actors, and actor of interest (indicated in yellow); (b) rasterized surroundings of the actor of interest (colored red) in bird's-eye view used as an input to CNN; (c) raster with overlaid ground-truth (dotted green line) and predicted (dotted blue line) 3s-trajectories}
\label{fig:example_output}
\vspace{-0.27cm}
\end{figure*}

To safely deploy SDVs to public roads one must solve a sequence of tasks that include detection and tracking of actors in SDV's surroundings, predicting their future trajectories, as well as navigating the SDV safely and effectively towards its intended destination while taking into account current and future states of the actors. We focus on a critical component of this pipeline, predicting future trajectories of tracked vehicles (in the following we use {\it vehicle} and {\it actor} interchangeably), where a working detection and tracking system is assumed. Our main contributions are as follows:
\begin{itemize}
  \item We propose to rasterize high-definition maps and surroundings of each vehicle in SDV's vicinity, thus providing complete context and information necessary for accurate prediction of future trajectory;
  \item We trained deep convolutional neural network (CNN) to predict short-term vehicle trajectories, while accounting for inherent uncertainty of motion in road traffic;
  \item Large-scale evaluation on real-world data showed that the system provides accurate predictions and well-calibrated uncertainties, indicating its practical benefits;
  \item Following extensive offline testing, the system was successfully tested onboard self-driving vehicles.
\end{itemize}
Example of a complex scene is shown in Figure \ref{fig:example_output}, where Fig. \ref{fig:example_output}a shows the scene in our internal 3D viewer, Fig. \ref{fig:example_output}b shows the rasterized 2D image (or {\it raster}) used as a model input, while Fig. \ref{fig:example_output}c shows 3-second ground-truth and predicted trajectories. Actor whose context corresponds to the raster is referred to as {\it actor of interest}. We can see that the method uses rasterization of surrounding map and actors to accurately predict actor movement in a dynamic environment.

\section{Related work}
In the past decade a number of methods were proposed to predict future motion of traffic actors. Comprehensive overview of the topic can be found in \cite{Lefevre2014,Wiest_2017}. Here, we review literature from the perspective of autonomous driving domain. We first cover engineered approaches commonly used in practice. Then, we discuss learned approaches using classical machine learning as well as deep learning methods.

\subsection{Motion prediction in self-driving systems}
Accurate prediction of actor motion is a critical component of deployed self-driving systems \cite{CosgunMCHDALTA17,Bertha2015}. In particular, prediction is tightly coupled with SDV's egomotion planning, as it is essential to accurately estimate future world state to correctly and safely plan for SDV's path through a highly dynamic environment. Inaccurate motion prediction may lead to severe accidents, as exemplified by a collision between MIT's ``Talos'' and Cornell's ``Skyne'' vehicles during the 2007 DARPA Urban Challenge \cite{Fletcher2009}.

Most of the deployed self-driving systems use well-established engineered approaches for motion prediction. The common approach consists of computing object's future motion by propagating its state over time based on kinematic models and assumptions of an underlying physical system. State estimate usually comprises position, speed, acceleration, and object heading, and techniques such as Kalman filter (KF) \cite{Kalman1960} are used to estimate and propagate the state in the future. For example, in Honda's deployed system \cite{CosgunMCHDALTA17}, KF tracker is used to predict motion of vehicles around SDV. While this approach works well for short-term predictions, its performance degrades for longer horizons as the model ignores surrounding context (e.g., roads, other traffic actors, traffic rules), as we confirm in Section \ref{sect:exp}. On the other hand, Mercedes-Benz's motion prediction component uses map information as a constraint to compute vehicle's future position \cite{Bertha2015}. The system first associates each detected vehicle with one or more lanes from the map. Then, all possible paths are generated for each {\it (vehicle, associated lane)} pair based on map topology, lane connectivity, and vehicle's current state. This heuristic provides reasonable predictions in most cases (as evaluated in Section \ref{sect:exp}), however it does not scale well nor is able to model unusual scenarios. As an alternative to existing deployed engineered approaches, by considering large amounts of data our proposed approach automatically learns that vehicles usually obey road and lane constraints, while also being capable of handling outliers.


\subsection{Learned prediction models}
Manually designed engineered models often impose unrealistic assumptions not supported by the data (e.g., that traffic always follows lanes), which motivated use of learned models as an alternative. 
A large class of learned models are maneuver-based models (e.g., using Hidden Markov Model \cite{Streubel2014}) which are object-centric approaches that predict discrete action of each object independently. The independence assumption does not often hold true, which is mitigated by the use of Bayesian networks \cite{Schreier2016} that are computationally more expensive and not feasible in real-time tasks. Additionally, in \cite{Ballan2016} authors learned scene-specific motion patterns and applied them to novel scenes with an image-based similarity function. However, these methods also require manually designed features to capture context information, resulting in suboptimal performance. 
Alternatively, Gaussian Process (GP) regression can be used to address the motion prediction problem \cite{Wang2008}. GP regression is well-suited for the task with desirable properties such as ability to quantify uncertainty, yet it is limited when modeling complex actor-environment interactions. In recent work researchers focused on how to model environmental context using Inverse Reinforcement Learning (IRL) \cite{ng2000algorithms} approaches. Kitani {\it et al.} \cite{Kitani2012} used inverse optimal control to predict pedestrian paths by considering scene semantics, however the proposed IRL methods are inefficient for real-time applications. 

The success of deep learning \cite{Goodfellow2016} motivated its use in the self-driving domain. In \cite{bojarski2016end} an end-to-end system that directly maps input sensors to SDV controls was proposed. In \cite{Lee2017} the authors described a Recurrent Neural Network (RNN)-based method for long-term predictions of interacting agents given scene context. In \cite{Alahi2016} authors proposed a social Long Short-Term Memory (LSTM) to model human movement together with social interactions. Authors of \cite{Fragkiadaki2016} used LSTM to predict ball motion in billiards directly from images. In \cite{wang2016classifying} LSTM models were used to classify basketball plays, with overhead raster images taken as inputs. Similarly, the authors of \cite{ondruska2016end,ondruvska2016deep} used overhead rasters and RNNs to track multiple objects in a scene by predicting raster image in a next timestep, unlike our work where full per-object trajectories are directly inferred. 
Due to strict time constraints of an onboard real-time system and the requirement to more easily debug and understand model decisions made on public roads, in this work we used simpler feed-forward CNN architectures for the prediction task. In addition, recent work indicates temporal CNNs could be more powerful than RNNs \cite{lea2017temporal}, further justifying our choice.

A critical feature for the safety of SDVs is uncertainty estimation for predictions. We address this important issue in our current work, building on an existing body of literature. This includes \cite{Alahi2016}, where the authors estimate uncertainty due to observation noise (i.e., {\it aleatoric} uncertainty) by learning to predict the parameters of assumed noise distribution. The authors of \cite{gal2016uncertainty} showed that dropout training in deep networks approximates uncertainty of the prediction model itself (i.e., {\it epistemic} uncertainty). In a followup work, \cite{kendall2017uncertainties} presented a deep method that jointly estimates aleatoric and epistemic uncertainties. Some recent publications have addressed uncertainty estimation in motion prediction from a self-driving perspective. For example, \cite{Bhattacharyya_2018} models both aleatoric and epistemic uncertainties of pedestrian and bicyclist motion over a 1-second horizon. Authors of \cite{bhattacharyya2019bayesian} developed a novel optimization scheme for dropout-based Bayesian inference using synthetic likelihoods to accurately capture model uncertainty. Lastly, \cite{huang2019uncertaintyaware} generated conditional variational distribution of predicted trajectories together with confidence estimates for different horizons. However, in contrast to our work, the proposed approach does not utilize high-definition maps and assumes that observation sensors are present on the actor of interest.

\section{Proposed approach}

Let us assume that we have access to real-time data streams coming from sensors such as lidar, radar, or camera, installed aboard a self-driving vehicle. Furthermore, we assume to have an already functioning tracking system ingesting the sensor data, allowing detection and tracking of traffic actors in real-time. For example, we can make use of any of a number of Kalman filter-based methods that have found wide practical use \cite{chen2012kalman}, taking sensor data as input and outputting tracks of individual actors that represent their state estimates at fixed intervals. State estimates contain the following information describing an actor: bounding box, position, velocity, acceleration, heading, and heading change rate. Lastly, we assume access to mapping data of an operating area, comprising road and crosswalk locations, lane directions, and other relevant map information. 

Let us denote high-definition map data by $\mathcal{M}$, and a set of discrete times at which tracker outputs state estimates as $\mathcal{T} = \{t_1, \dots, t_T\}$, where time gap between consecutive time steps is constant (e.g., gap is equal to $0.1s$ for tracker running at the frequency of $10Hz$). Then, we denote state output of a tracker for the $i$-th actor at time $t_j$ as ${\bf s}_{ij}$, where $i = 1, \dots, N_j$ with $N_j$ being a number of unique actors tracked at time $t_j$. Note that in general actor counts vary for different time steps as new actors appear within and existing ones disappear from the sensor range. Then, given data $\mathcal{M}$ and all actors' state estimates up to and including time step $t_j$ (denoted by $\mathcal{S}_j$), the task is to predict sequence of future states $[{\bf s}_{i(j+1)}, \dots, {\bf s}_{i(j+H)}]$, where $H$ denotes the number of future consecutive time steps for which we predict states (or {\it prediction horizon}). Without the loss of generality, we simplify the task to infer $i$-th actor's future positions instead of full state estimates, denoted as $[x_{i(j+1)}, \dots, x_{i(j+H)}]$ for $x$- and similarly for $y$-positions. Past and future positions at time $t_j$ are represented in actor-centric coordinate system derived from actor's state at time $t_j$, where forward direction represents $x$-axis, left-hand direction represents $y$-axis, and actor's bounding box centroid represents the origin.

\subsection{Model inputs}
\label{sect:input_format}


To model dynamic context at time $t_j$ we use state data $\mathcal{S}_j$, while to model static context we use map data $\mathcal{M}$, comprising road and crosswalk polygons, as well as lane directions and boundaries. Road polygons describe drivable surface, lanes describe driving path, and crosswalk polygons describe road surface used for pedestrian crossing. Lanes are encoded by boundaries and directed lines positioned at the center.

Instead of manually defining features that represent actor context, we propose to rasterize a scene for the $i$-th actor at time step $t_j$ into an RGB image (see Figure \ref{fig:example_output} for an example). Then, using rasterized images as inputs we train CNN to predict actor trajectory, where the network automatically infers relevant features. Optionally, the model can also take as input a current state of the actor of interest ${\bf s}_{ij}$ represented as a vector (see Section \ref{sect:network_diag} for details of the architecture).

\subsubsection{Rasterization}
\label{sect:data_representation}
To describe rasterization, let us first introduce a concept of a {\it vector layer}, formed by a collection of polygons and lines that belong to a common type. For example, in the case of map elements we have vector layer of roads, of crosswalks, and so on. To rasterize vector layer into an RGB space, each vector layer is manually assigned a color from a set of distinct RGB colors that make a difference among layers more prominent. The only layer that does not have its defined RGB color is a layer that encodes lane direction. Instead of assigning a specific RGB color, we use a direction of each straight line segment as a hue value in HSV color space \cite{Smith1978}, with saturation and value set to maximum. 
The hue component is angular measurement and corresponds to a position at a color wheel, with hue of $0^{\circ}$ indicating red, $120^{\circ}$ indicating green, and blue corresponding to $240^{\circ}$. 
We then convert HSV to RGB color space, thus encoding driving direction of each lane in the resulting raster image. For example, in Figure \ref{fig:example_output} lanes going in opposite directions are represented by colors diametrically opposite to each other on the HSV color cylinder. Once the colors are defined, vector layers are rasterized one by one on top of each other, in the order from layers that represent larger areas such as road polygons towards layers that represent finer structures such as lanes or actor bounding boxes. Important parameter is pixel resolution, which we set to $0.1m$ considering trade-off between image size and ability to represent fine details.

As discussed earlier, we are interested in representing context for each actor separately. To represent context around the $i$-th actor tracked at time step $t_j$ we create a rasterized image $I_{ij}$ of size $n \times n$ such that the actor is positioned at pixel $(w, h)$ within $I_{ij}$, where $w$ represents width and $h$ height measured from the bottom-left corner of the image. The image is rotated such that actor's heading points up, where lane directions are computed relative to the actor's heading and then encoded in the HSV space. 
We set $n = 300$, actor of interest is positioned at $w = 150$ and $h = 50$, so that $25m$ in front of the actor and $5m$ from the back is rasterized (for our experiments we only considered roads with maximum speed limit of $25mph$ where this setup performs well, for faster roads more context would be required). Lastly, we color the actor of interest differently so that it is distinguishable from other surrounding vehicles (as seen in Figure \ref{fig:example_output}b the actor of interest is colored red, while all others are colored yellow).

To capture past motion of all traffic actors, their bounding boxes at consecutive time steps $[t_{j-K+1}, \dots, t_{j}]$ are rasterized on top of map vector layers. Each historical actor polygon is rasterized with the same color as the current polygon yet with reduced level of brightness, resulting in the {\it fading} effect. Brightness level at $t_{j-k}$ is equal to $\max(0, 1 - k \cdot \delta)$, $k = 0, 1, \dots, K-1$, where we set $\delta = 0.1$ and $K$ to either $1$ (no fading) or $5$ (with fading, example shown in Figure \ref{fig:example_output}b). 

Note that we consider map data and tracked states of all traffic actors to generate rasters, and do not use raw sensor data (i.e., camera, lidar, or radar) for rasterization. Moreover, although we did not observe a significant effect for different color selections, we recognize that the rasterization could be further optimized. For example, layer ordering can be modified, along with the raster size, resolution, and other parameters. However, due to limited space this is outside of the scope of the current work, and in the following we use the stated parameter values found to work well in practice.


\begin{figure*}
\centering
\begin{minipage}{.71\textwidth}
  \centering
\includegraphics[keepaspectratio=1.0,width=0.85\columnwidth]{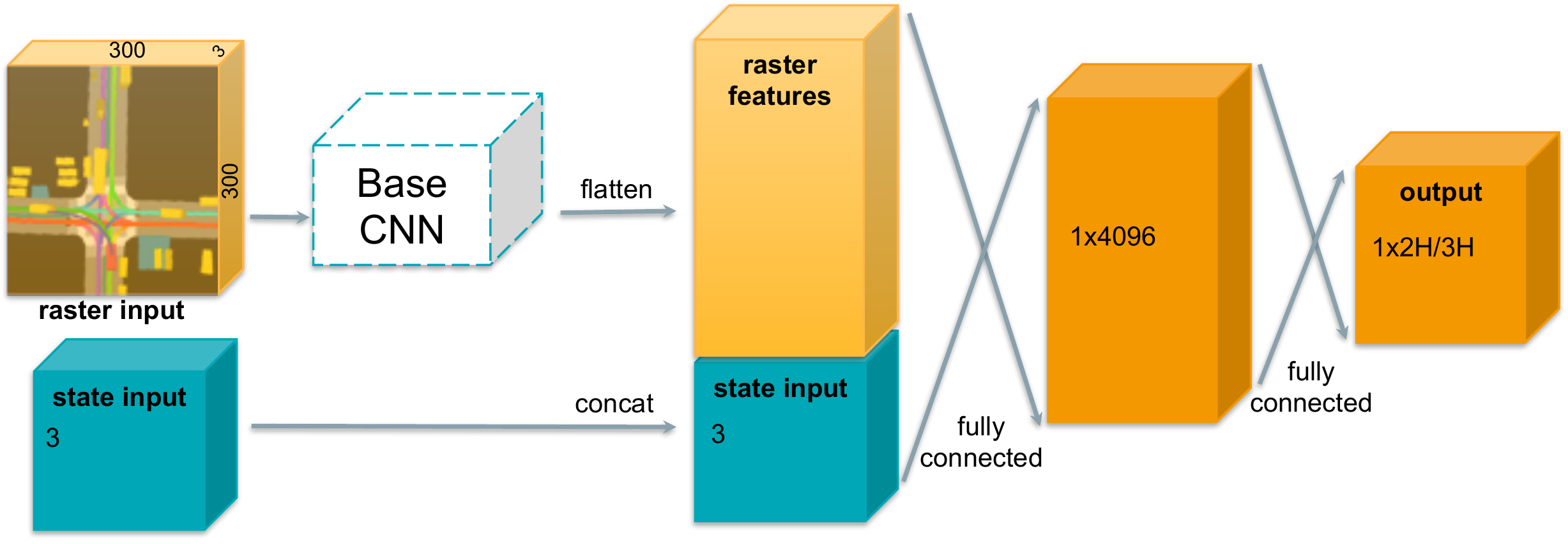}
\caption{Feed-forward network architecture combining raster image and actor state inputs}
\label{fig:network_diag}
\end{minipage}%
\begin{minipage}{.29\textwidth}
\includegraphics[keepaspectratio=1.0,width=1.0\columnwidth]{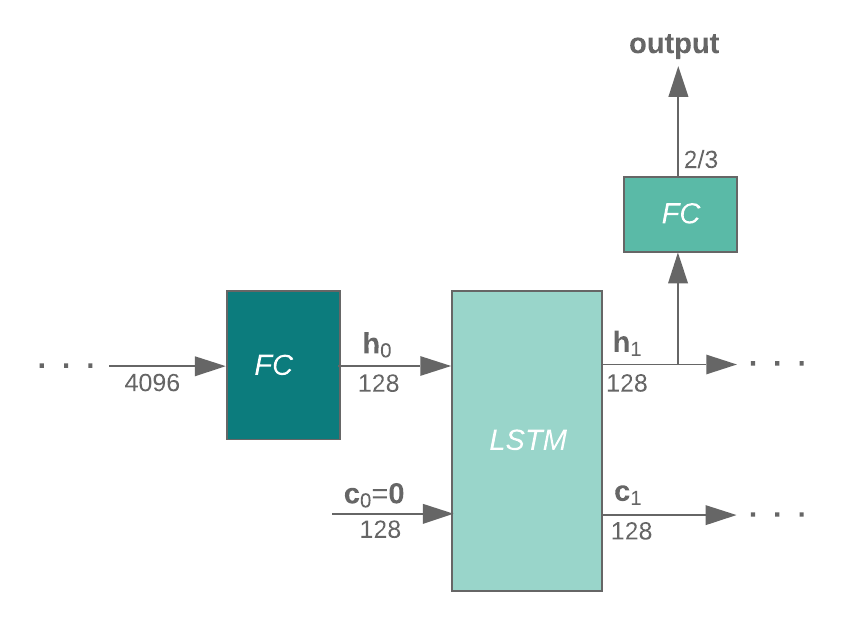}
\caption{LSTM decoder}
\label{fig:lstm_decoder}
\end{minipage}
\vspace{-.15in}
\end{figure*}

\subsection{Optimization problem}
To obtain analytical expressions for loss functions used to optimize deep networks, let us first introduce {\it displacement error} for the $i$-th actor at time $t_j$ for horizon $h \in \{1, \dots, H\}$,
\begin{equation}
\begin{split}
d_{i(j+h)} = & \Big(\big(x_{i(j+h)} - {\hat x}_{i(j+h)}(\mathcal{S}_j, \mathcal{M}, \theta)\big)^2 + \\ & \big(y_{i(j+h)} - {\hat y}_{i(j+h)}(\mathcal{S}_j, \mathcal{M}, \theta)\big)^2\Big)^{1/2},
\end{split}
\end{equation}
defined as Euclidean distance between observed and predicted positions. Here, $\theta$ denotes parameters of a model, while ${\hat x}_{i(j+h)}(\mathcal{S}_j, \mathcal{M}, \theta)$ and ${\hat y}_{i(j+h)}(\mathcal{S}_j, \mathcal{M}, \theta)$ denote position outputs of the model that takes available states $\mathcal{S}_j$ and map $\mathcal{M}$ as inputs. Then, overall loss incurred by predicting trajectory for a complete prediction horizon is equal to average squared displacement error of trajectory points,
\begin{equation}
\label{eq:loss_simple}
L_{ij} = \frac{1}{H} \sum_{h=1}^H d_{i(j+h)}^2,
\end{equation}
where we train the model to output $2H$-D vector, representing predicted $x$- and $y$-positions for each of $H$ trajectory points. Optimizing over all actors and time steps, we find optimal parameters by minimizing overall training loss,
\begin{equation}
\label{eq:overall_loss}
\theta^* = \argmin_{\theta} \mathcal{L} = \argmin_{\theta} \sum_{j=1}^T \sum_{i=1}^{N_j} L_{ij}.
\end{equation}

Alternatively, as the prediction task is inherently noisy it is useful to capture aleatoric uncertainty present in the data \cite{kendall2017uncertainties,lakshminarayanan2017simple}, in addition to optimizing for a point estimate as in (\ref{eq:overall_loss}). To that end, we assume that displacement errors are sampled from a half-normal distribution \cite{johnson1962folded}, denoted as
\begin{equation}
d_{i(j+h)} \sim \mathcal{FN}\big(0, {\hat \sigma}_{i(j+h)}(\mathcal{S}_j, \mathcal{M}, \theta)^2\big),
\end{equation}
where standard deviation ${\hat \sigma}_{i(j+h)}$ is computed by the model. Then, we can write overall loss for the $i$-th actor at time $t_j$ as negative log-likelihood of the observed data, equal to
\begin{equation}
\label{eq:uncertain_loss}
L_{ij} = \sum_{h=1}^H \Big( \frac{d_{i(j+h)}^{2}}{2 ~{\hat \sigma}_{i(j+h)}(\mathcal{S}_j, \mathcal{M}, \theta)^2} + \log {\hat \sigma}_{i(j+h)}(\mathcal{S}_j, \mathcal{M}, \theta) \Big),
\end{equation}
where we train the model to output $3H$-dimensional vector, representing predicted $x$- and $y$-positions, as well as standard deviation for $H$ trajectory points. Lastly, optimizing over entire training data we solve (\ref{eq:overall_loss}) with $L_{ij}$ computed as in (\ref{eq:uncertain_loss}).

\subsection{Network architecture}
\label{sect:network_diag}
In this section we describe an architecture used to solve the optimization problems \eqref{eq:loss_simple} and \eqref{eq:uncertain_loss}, also illustrated in Figures \ref{fig:network_diag} and \ref{fig:lstm_decoder}. To extract features from an input raster we can use any existing CNN (referred to as {\it base CNN}). In addition, to input actor state we encode it as a 3D vector comprising velocity, acceleration, and heading change rate (position and heading are not required as they were already used during raster generation), and concatenate the resulting vector with flattened output of the base CNN. Then, the combined features are passed through a fully-connected (FC) layer (we set its size to $4{,}096$) connected to an output layer of size $2H$ if solving (\ref{eq:loss_simple}), or $3H$ if solving (\ref{eq:uncertain_loss}). 

Alternatively, we can decode the actor trajectory through a recurrent architecture, using an LSTM \cite{hochreiter1997long} after the first FC layer (shown in Figure \ref{fig:lstm_decoder}). We set LSTM size to $128$, cell state is 0-initialized, while initial input is obtained by converting output of the FC layer of size $4{,}096$ into a vector of size $128$ with another FC layer. For each time step LSTM output is converted by an output FC layer into a $2$-D vector if solving (\ref{eq:loss_simple}) or a $3$-D vector if solving (\ref{eq:uncertain_loss}) (representing $x$- and $y$-position, and standard deviation).

\section{Experiments}
\label{sect:exp}
In this section we present detailed results of empirical evaluation of the proposed deep convolutional approach. 


{\bf Data}
We collected 240 hours of data by manually driving SDV in Pittsburgh, PA and Phoenix, AZ in various traffic conditions (e.g., varying times of day, days of the week), with data collection rate of $10Hz$. We ran a state-of-the-art detector and Unscented KF (UKF) tracker \cite{wan2000unscented} with the kinematic state-transition model \cite{kong2015kinematic} on this data to produce a set of tracked vehicle detections.
Each tracked actor at each discrete tracking time step amounts to one data point, with overall data comprising $7.8$ million examples after removing static actors. We considered prediction horizon of $3s$ (i.e., we set $H = 30$), and used 3:1:1 split to obtain train/validation/test data.

{\bf Baselines}
1) We used UKF to predict future motion by forward propagating estimated states in time.
2) We used a linear baseline that directly converts input states (of size $3$) into future positions for each time step.
3) Vehicle-lane association \cite{Bertha2015} that considers map constraints was used. More specifically, an actor was assigned to nearby lanes within $5m$ radius, and Pure Pursuit algorithm \cite{coulter1992implementation} with dynamic lookahead \cite{chen1999experimental} was used to follow that lane. If there are multiple associated lanes, the one with the lowest error was reported (denoted as {\it lane-assoc}).

{\bf Models}
We compared the baselines to several variants of the proposed approach. We considered the following base CNNs: AlexNet \cite{alexnet}, VGG-19~\cite{vgg}, ResNet-50~\cite{resnet}, and MobileNet-v2 (MNv2) \cite{sandler2018inverted}.
Furthermore, to evaluate how varying input complexity affects the performance, we considered architectures that use: 1) raster without fading and state, solving (\ref{eq:loss_simple}); 2) raster with fading and without state, solving (\ref{eq:loss_simple}); 3) raster without fading and with state, solving (\ref{eq:loss_simple}); 4) raster with fading and state, solving (\ref{eq:loss_simple}); 5) raster with fading and state, and outputting uncertainty, solving (\ref{eq:uncertain_loss}).

\begin{table*} [!ht]
\small
\caption{Comparison of average prediction errors for competing methods (in meters)}
\label{tab:pred-errors}
\centering
{
  \begin{tabular}{lcccccc}
    {\bf Method} & {\bf Raster} & {\bf State} & {\bf Loss} & {\bf Displacement} & {\bf Along-track} & {\bf Cross-track} \\
    \hline
    \rowcolor{lightgray}
    UKF & -- & yes & -- & 1.46 & 1.21 & 0.57 \\
    Linear model & -- & yes & (\ref{eq:loss_simple}) & 1.19 & 1.03 & 0.43 \\
    \rowcolor{lightgray}
    Lane-assoc & -- & yes & -- & 1.09 & 1.09 & 0.19 \\
    \hline
    AlexNet & w/o fading & no & (\ref{eq:loss_simple}) & 3.14 & 3.11 & 0.35 \\
    \rowcolor{lightgray}
    AlexNet  & w/ fading & no & (\ref{eq:loss_simple}) & 1.24 & 1.23 & 0.22 \\
    AlexNet  & w/o fading & yes & (\ref{eq:loss_simple}) & 0.97 & 0.94 & 0.21 \\
    \rowcolor{lightgray}
    AlexNet  & w/ fading & yes & (\ref{eq:loss_simple}) & 0.86 & 0.83 & 0.20 \\
    \hline    VGG-19  & w/ fading & yes  & (\ref{eq:loss_simple}) & 0.77 & 0.75 & 0.19 \\
    \rowcolor{lightgray}
    ResNet-50 & w/ fading & yes  & (\ref{eq:loss_simple}) & 0.76 & 0.74 & {0.18} \\
    MobileNet-v2 & w/ fading & yes  & (\ref{eq:loss_simple}) & {0.73} & {0.70} & {0.18} \\
    \hline
    \rowcolor{lightgray}
    MobileNet-v2 & w/ fading & yes  & (\ref{eq:uncertain_loss}) & {0.71} & {0.68} & {0.18} \\
    MobileNet-v2 LSTM & w/ fading & yes  & (\ref{eq:uncertain_loss}) & {\bf 0.62} & {\bf 0.60} & {\bf 0.14} \\
    \hline
\end{tabular}
}
\end{table*}

{\bf Training} Models were implemented in TensorFlow \cite{tensorflow2015-whitepaper} and trained on 16 Nvidia Titan X GPU cards. To coordinate the GPUs we used open-source framework Horovod \cite{sergeev2018horovod}, completing training in around 24 hours. We used per-GPU batch size of $64$ and trained with Adam optimizer \cite{kingma2014adam}, setting the initial learning rate to $10^{-4}$ that was further decreased by a factor of $0.9$ every 20 thousand iterations. All models were trained end-to-end from scratch, except for a model with uncertainty outputs which was initialized with a corresponding model without uncertainty and then fine-tuned (training from scratch did not give satisfactory results). 

\begin{figure}[!t]
\centering
\includegraphics[keepaspectratio=1,width=0.49\columnwidth]{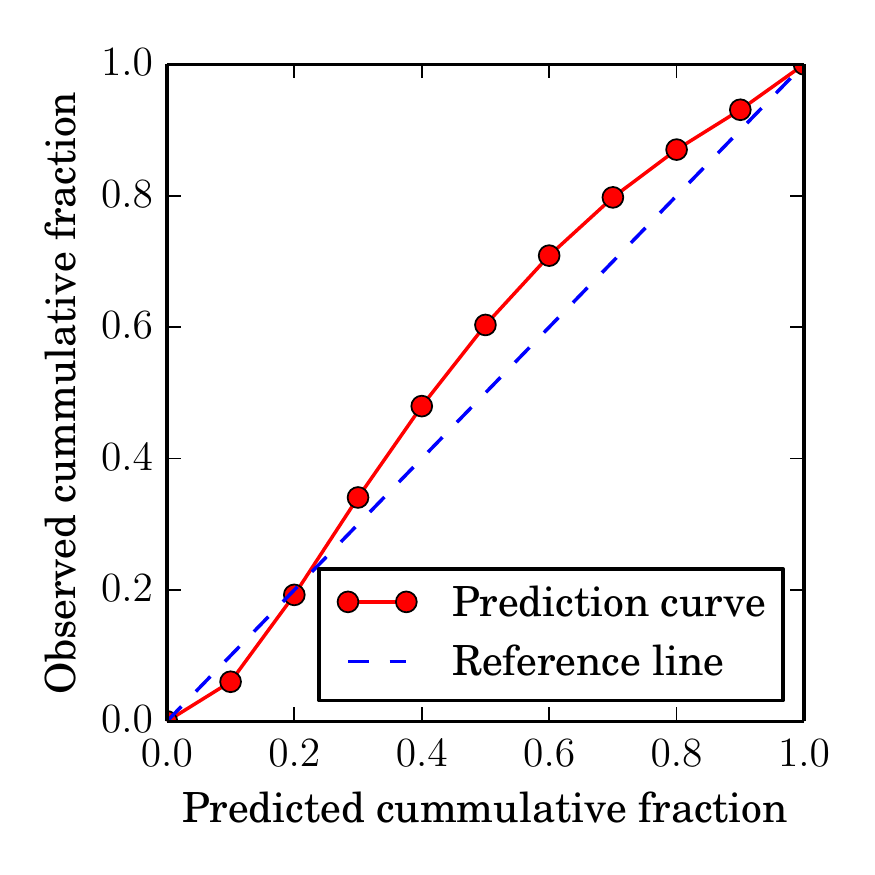}
\includegraphics[keepaspectratio=1,width=0.49\columnwidth]{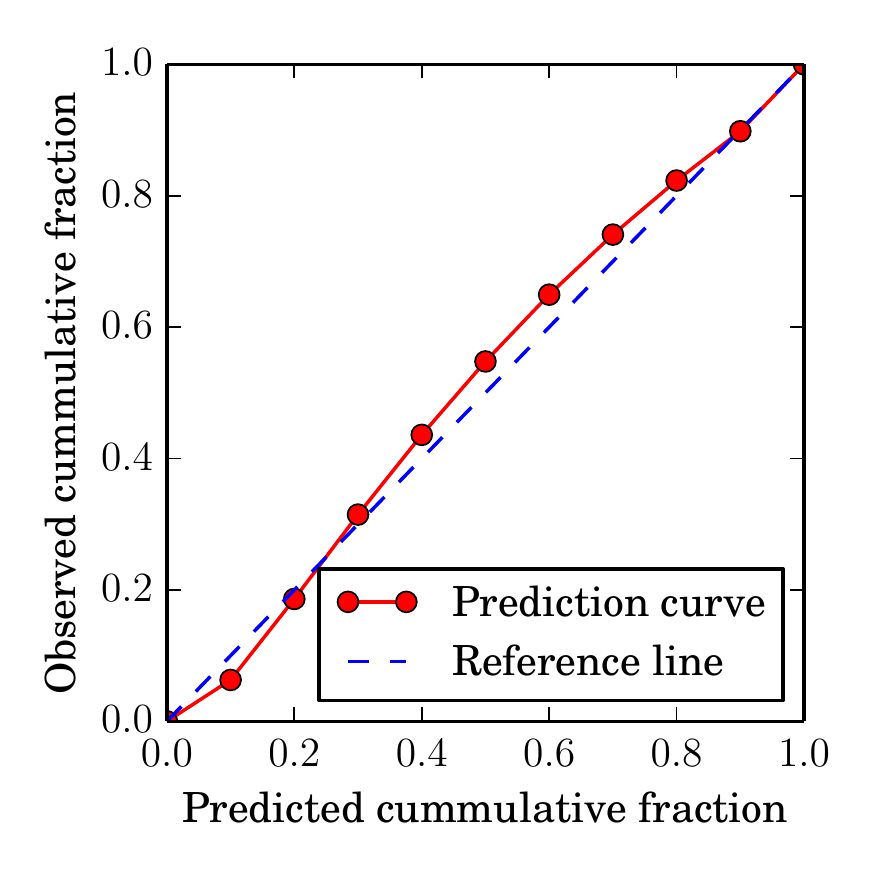}
\caption{Reliability diagrams at horizons of: (a) 1s; (b) 3s}
\label{fig:reliability_diagrams}
\end{figure}

\begin{figure*}[!th]
\centering
\subfigure{
\includegraphics[keepaspectratio=1,width=0.48\columnwidth,trim={0 1.3cm 0cm 0cm},clip]{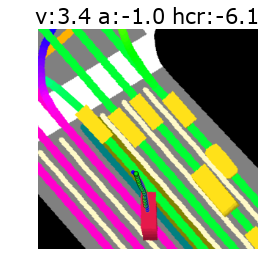}
\label{fig:traj_pred_examples_a}
}
\subfigure{
\includegraphics[keepaspectratio=1,width=0.48\columnwidth,trim={0 1cm 0 0.7cm},clip]{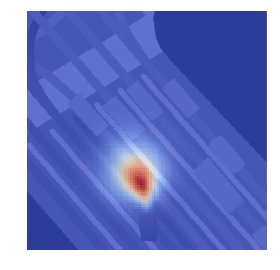}
\label{fig:traj_pred_examples_a}
}
\subfigure{
\includegraphics[keepaspectratio=1,width=0.48\columnwidth,trim={0 1cm 0 0.7cm},clip]{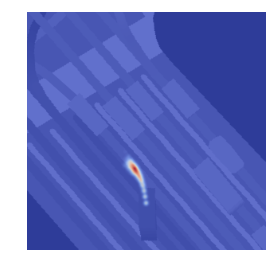}
\label{fig:traj_pred_examples_a}
}
\subfigure{
\includegraphics[keepaspectratio=1,width=0.48\columnwidth,trim={0 1cm 0 0.7cm},clip]{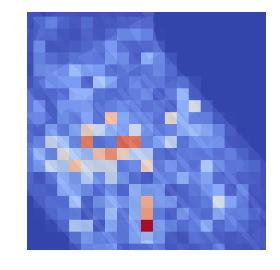}
\label{fig:traj_pred_examples_a}
}
\\
\subfigure{
\includegraphics[keepaspectratio=1,width=0.48\columnwidth,trim={0 1.3cm 0 0.0cm},clip]{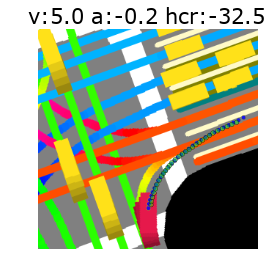}
\label{fig:traj_pred_examples_a}
}
\subfigure{
\includegraphics[keepaspectratio=1,width=0.48\columnwidth,trim={0 1cm 0 0.7cm},clip]{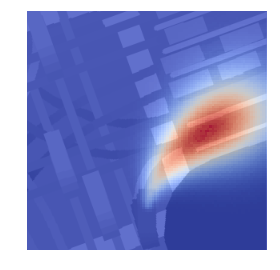}
\label{fig:traj_pred_examples_a}
}
\subfigure{
\includegraphics[keepaspectratio=1,width=0.48\columnwidth,trim={0 1cm 0 0.7cm},clip]{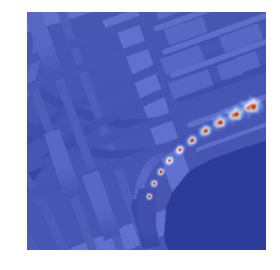}
\label{fig:traj_pred_examples_a}
}
\subfigure{
\includegraphics[keepaspectratio=1,width=0.48\columnwidth,trim={0 1cm 0 0.7cm},clip]{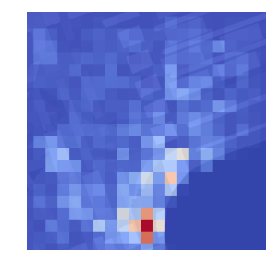}
\label{fig:traj_pred_examples_a}
}
\\
\subfigure{
\includegraphics[keepaspectratio=1,width=0.48\columnwidth,trim={0 1.3cm 0 0.0cm},clip]{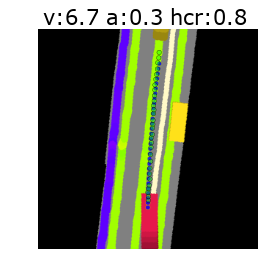}
\label{fig:traj_pred_examples_a}
}
\subfigure{
\includegraphics[keepaspectratio=1,width=0.48\columnwidth,trim={0 1cm 0 0.7cm},clip]{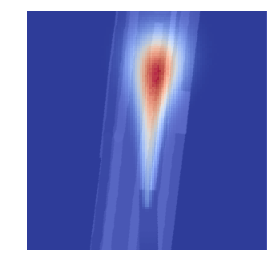}
\label{fig:traj_pred_examples_a}
}
\subfigure{
\includegraphics[keepaspectratio=1,width=0.48\columnwidth,trim={0 1cm 0 0.7cm},clip]{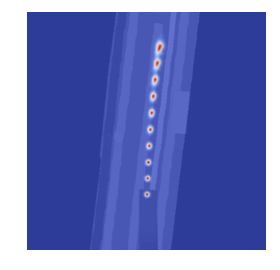}
\label{fig:traj_pred_examples_a}
}
\subfigure{
\includegraphics[keepaspectratio=1,width=0.48\columnwidth,trim={0 1cm 0 0.7cm},clip]{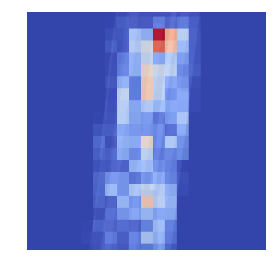}
\label{fig:traj_pred_examples_a}
}
\caption{Analysis of the MNv2 model on the three case studies, with results overlaid over the input raster images; the first column shows ground-truth (dotted green line) and predicted (dotted blue line) 3-second trajectories, the second column shows aleatoric uncertainty output by the model, the third column shows epistemic uncertainty estimated by dropout analysis, the fourth column shows relevant parts of raster estimated by occlusion sensitivity analysis; state inputs are provided above the rasters in the first column, indicating velocity (v) in $m/s$, acceleration (a) in $m/s^2$, heading change rate (hcr) in $deg/s$}
\vspace{-0.2cm}
\label{fig:traj_pred_examples}
\end{figure*}

\subsection{Results}
In Table \ref{tab:pred-errors} we report error metrics relevant for motion prediction: displacement errors, as well as along-track and cross-track errors \cite{gong2004methodology}, averaged over the prediction horizon. We emphasize that metrics improvements of even a couple of centimeters can make a large difference in practice, significantly affecting the safety and comfort of SDVs.

Considering the baselines, we see that the linear model easily outperformed the baseline UKF, which simply propagates an initial actor state. Moreover, using the map information through the lane-assoc model we gained significant improvements, especially in the cross-track which is already at the level of the best deep models. This is an expected result, as vehicles usually follow their lanes quite well.

We then conducted an ablation study using the feed-forward architecture from Figure \ref{fig:network_diag} and AlexNet as a base CNN, running experiments with varying input complexity (upper half of Table \ref{tab:pred-errors}).
When we provide neither fading nor state inputs the model performs worse than UKF, as the network does not have enough information to estimate current state of an actor from the raster. Interestingly, when we include fading the model starts to outperform the baseline by a large margin, indicating that actor state can be inferred solely from providing past positions through fading. If instead of fading we directly provide state estimates we get even better performance, as the state info is already distilled and does not need to be estimated from raster. Furthermore, using raster with fading together with state inputs leads to additional performance boost, suggesting that fading carries additional info not available through the state itself, and that the raster and other external inputs can be seamlessly combined through the proposed architecture to improve accuracy. 

Next, we compared popular CNN architectures as base CNNs. As seen in the bottom half of Table \ref{tab:pred-errors}, we found that VGG and ResNet models provide improvements over the baseline AlexNet, as observed previously \cite{vgg}. It is interesting to note that only starting with these models did we outperform the baseline lane-assoc model in terms of all the relevant metrics. However, both models are outperformed by the novel MNv2 architecture that combines a number of deep learning ideas under one roof (e.g., bottleneck layers, residual connections, depthwise convolutions). 
Taking the best performing MNv2 as a base and extending the output layer by adding uncertainty led to further improvements. Not only do additional outputs allow estimation of trajectory uncertainty in addition to trajectory point estimates, but they also mitigate adverse effects of noisy data during the training process. Lastly, using LSTM decoder at the output, as described in Section \ref{sect:network_diag}, led to the best results. In our task the future states depend on the past ones, which can be captured by the recurrent architecture. 
In the remainder we analyze results of the state-of-the-art MNv2 model in greater detail.

We used reliability diagrams to evaluate how closely predicted error distribution matches testing error distribution. The diagrams are generated by measuring how large is an observed displacement error compared to a predicted confidence, and computing what fraction of observed errors falls within the expected range given by the estimated standard deviation. For example, due to the Gaussianity assumption we expect $68\%$ of observed errors to be within the predicted one sigma, and diagram point at predicted value of $0.68$ should be as close as possible to observed value of $0.68$. Thus, the closer the curve is to the diagonal line, the better calibrated is the model.
Figure~\ref{fig:reliability_diagrams} shows diagrams for horizons of $1s$ and $3s$. The prediction curve is well aligned with the reference line, especially at $3$ seconds whereas $1s$-predictions are slightly underconfident. Thus, given an estimated sigma, we can expect with high confidence that in $68\%$ of cases an actual error will not be larger than that value. Plots for other horizons are omitted as they resemble the ones shown.

\subsection{Case studies}
In Figure \ref{fig:traj_pred_examples} we give example outputs for three scenes commonly encountered in traffic. As we will see, the model provided accurate short-term trajectories in all the cases, as well as reasonable and intuitive uncertainty estimates. 

The first case (first row) involves actor cutting over opposite lanes when entering road from off-street parking, where the model correctly predicted that the actor will queue for vehicles in front (image in the first column). The uncertainty estimates reflect peculiarity of the situation (image in the second column), as the actor is not following common traffic rules and may choose to either queue for the leftmost vehicle or cut the road to queue for the vehicles in the other lanes. In the second row we see an actor making a right turn in an intersection, where the model correctly predicted that the actor is planning to enter its own lane. However, uncertainty increases compared to the first example, as the vehicle has higher speed as well as heading change rate, and there is a possibility it may enter any of the two vacant lanes. Lastly, in the third row we have a fast actor going straight, while changing lanes to avoid an obstacle. The lane change is correctly predicted, as well as lower cross-track uncertainty due to actor's higher speed. Quite intuitively, probability that the actor hits the obstacle is estimated to be near-zero.

Next, we performed a dropout analysis to estimate uncertainty within the model itself (i.e., epistemic uncertainty) \cite{kendall2017uncertainties}, done by dropping out $50\%$ of randomly selected nodes in the fully-connected layers from Figure \ref{fig:network_diag}, repeating the process $100$ times, and visualizing variance of the resulting trajectory points. The results are shown in the third column of Figure \ref{fig:traj_pred_examples}, where we can see that the epistemic uncertainty is very low in all cases, in fact several orders of magnitude lower than aleatoric (or process) uncertainty visualized in the second column. This indicates that the model converged, that more data would have limited effect on performance, and that the overall uncertainty can be approximated by considering only the learned uncertainty. 

\begin{figure}[t!]
\centering
\includegraphics[keepaspectratio=1,width=0.5\columnwidth,trim={0 0.7cm 0 0},clip]{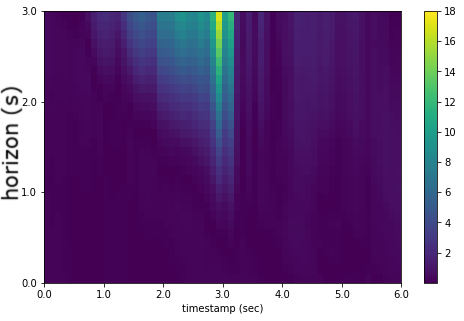}
\includegraphics[keepaspectratio=1,width=0.48\columnwidth,trim={0.7cm 0.7cm 0 0},clip]{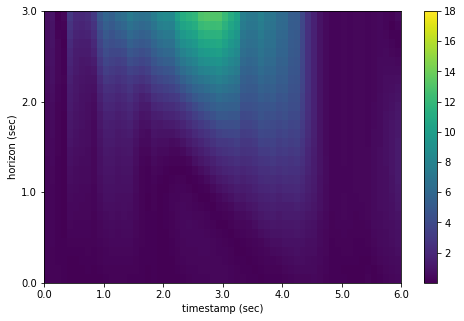}
\includegraphics[keepaspectratio=1,width=0.5\columnwidth]{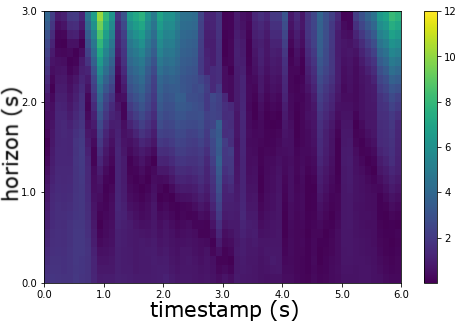}
\includegraphics[keepaspectratio=1,width=0.48\columnwidth,trim={0.7cm 0 0 0},clip]{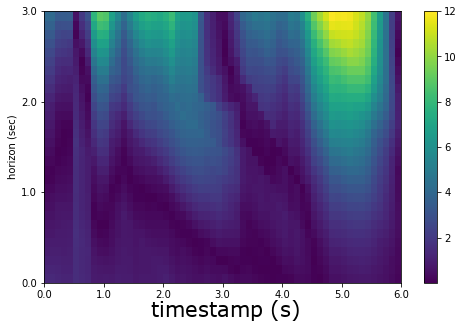}
\caption{Detailed analysis of cross- and along-track errors across various horizons for the second example shown in Figure \ref{fig:traj_pred_examples} (top: cross-track, bottom: along-track, left: MNv2 model, right: UKF model); $x$-axis indicates time of an event, $y$-axis indicates the prediction horizon, while color encodes an error in meters at each particular ({\it time}, {\it horizon}) pair}
\label{fig:error_heatmap}
\end{figure}

In addition, we performed sensitivity analysis \cite{zeiler2014visualizing} to understand which parts of the raster the model is focusing on. We swept a $15 \times 15$ black box across the raster and visualized the amount of change in the output compared to a non-occluded raster (as measured by the average displacement error), with results shown in the fourth column of Figure \ref{fig:traj_pred_examples}. In the first case the model focused on the oncoming lane and vehicles in front of the actor, as those parts of the raster are most relevant for a vehicle cutting across oncoming traffic and queuing. Quite intuitively, in the second case the model focused on nearby vehicles and crosswalks in the turn lane, while in the third case it focused on the obstacle and the lane further ahead due to actor's higher speed. Such analysis helps debug and understand what the model learned, and confirms it managed to extract knowledge from the training data that comes naturally to experienced human drivers.

In Figure \ref{fig:error_heatmap} we provide an additional analysis of cross- and along-track errors, using the second scenario from Figure \ref{fig:traj_pred_examples} as an example. At each timestamp of the event ($x$-axis), we color-code errors at each prediction horizon up to $3$ seconds in the future ($y$-axis). The actor starts to approach the intersection at around $1s$ mark, and initiates the turn at around $3s$ mark. Looking at the top two figures, we see that initially both MNv2 and UKF incorrectly predicted that the actor is going straight (note that allowed directions from the actor's current lane are straight and right), as indicated by the cross-track errors that are increasing as the prediction and the ground-truth started to diverge several seconds into the prediction horizon. However, we see that the proposed approach gave accurate prediction nearly at soon as the vehicle actually initiated its turn, and following $3.2s$ mark the cross-track errors dropped significantly. On the other hand, UKF took more time to catch up, and higher-error predictions lingered for nearly $1.5s$ more.
We see a similar situation when we compare along-track errors in the bottom two figures. The proposed approach consistently maintained lower error, which also dropped significantly when the actor started the turn. However, it is interesting to note that the error remained small even once the turn was complete (at around $5s$ mark), while UKF again required some time to capture the full actor state. We believe that such detailed analysis of individual cases, going beyond aggregated numbers and using the error heatmaps presented in Figure \ref{fig:error_heatmap}, could be useful to other researchers within the industry in their own work.

\begin{figure}[!t]
\centering
\includegraphics[keepaspectratio=1,width=1\columnwidth,trim={0 0.5cm 0 0},clip]{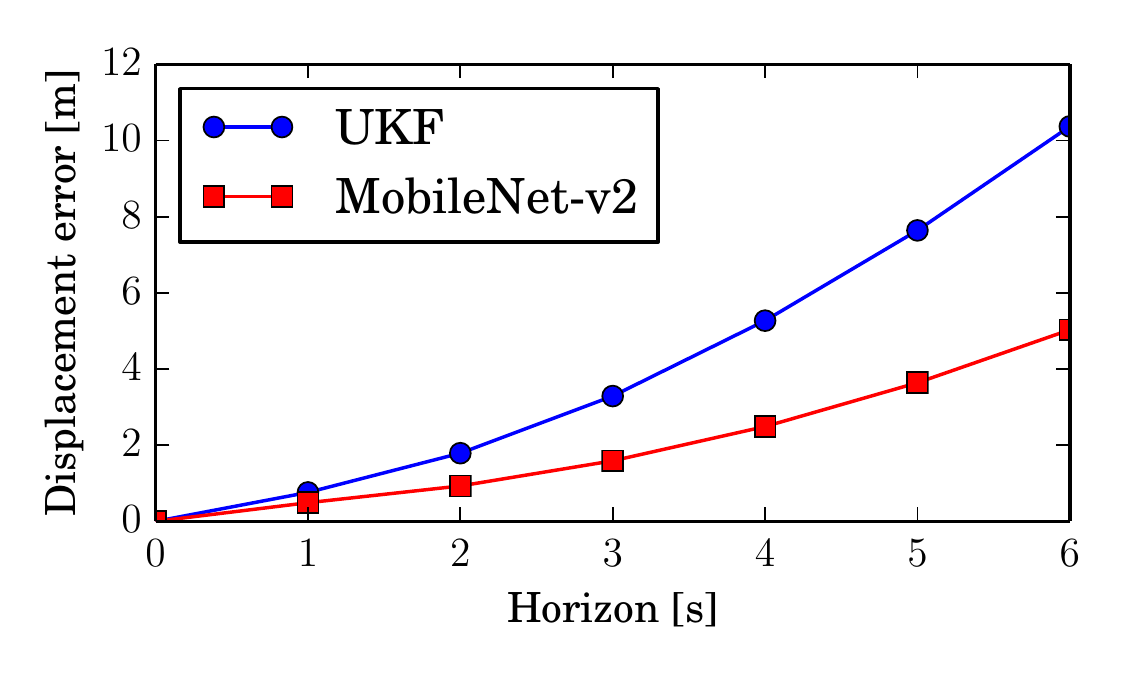}
\caption{Displacement error as a function of horizon}
\label{fig:error_vs_horizon}
\end{figure}

We are exploring several directions to improve the system. Most importantly, as the traffic domain is inherently multi-modal (e.g., actor approaching an intersection may turn left, right, or continue straight), we wanted to explore how far in the future does the proposed unimodal model provide useful predictions.
To answer this question we retrained a model with $H=60$ and measured performance at various horizons, with results given in Figure \ref{fig:error_vs_horizon}. While both UKF and the proposed method give reasonable short-term predictions, for longer horizons multimodality causes exponential error increase. To correctly model longer-term trajectories beyond the considered short-term $3s$ horizon we need to account for that aspect as well, which is a topic of our ongoing research.

\section{Conclusion}

We presented an effective solution to a critical part of the SDV problem, motion prediction of traffic actors. We introduced a deep learning-based method that provides both point estimates of future actor positions and their uncertainties. 
The method first rasterizes actor contexts, followed by training CNNs to use the resulting raster images to predict actor's short-term trajectory and the corresponding uncertainty. Extensive evaluation of the method strongly suggests its practical benefits, and following offline testing the framework was successfully tested onboard self-driving vehicles.



{
\bibliographystyle{ieee}
\bibliography{wacv2019}

\begin{thebibliography}{10}\itemsep=-1pt

\bibitem{tensorflow2015-whitepaper}
M.~Abadi, A.~Agarwal, P.~Barham, E.~Brevdo, et~al.
\newblock {TensorFlow}: Large-scale machine learning on heterogeneous systems,
  2015.

\bibitem{Alahi2016}
A.~Alahi, K.~Goel, et~al.
\newblock {\em Social LSTM: Human Trajectory Prediction in Crowded Spaces}.
\newblock IEEE, Jun 2016.

\bibitem{Ballan2016}
L.~Ballan, F.~Castaldo, A.~Alahi, F.~Palmieri, and S.~Savarese.
\newblock {\em Knowledge Transfer for Scene-Specific Motion Prediction}, page
  697–713.
\newblock Springer International Publishing, 2016.

\bibitem{Bhattacharyya_2018}
A.~Bhattacharyya, M.~Fritz, and B.~Schiele.
\newblock Long-term on-board prediction of people in traffic scenes under
  uncertainty.
\newblock {\em 2018 IEEE/CVF Conference on Computer Vision and Pattern
  Recognition}, Jun 2018.

\bibitem{bhattacharyya2019bayesian}
A.~Bhattacharyya, M.~Fritz, and B.~Schiele.
\newblock Bayesian prediction of future street scenes using synthetic
  likelihoods, 2019.

\bibitem{national2014economic}
L.~J. Blincoe, T.~R. Miller, E.~Zaloshnja, and B.~A. Lawrence.
\newblock The economic and societal impact of motor vehicle crashes, 2010
  (revised).
\newblock Technical Report DOT HS 812 013, National Highway Traffic Safety
  Administration, May 2015.

\bibitem{bojarski2016end}
M.~Bojarski, D.~Del~Testa, D.~Dworakowski, B.~Firner, B.~Flepp, P.~Goyal, L.~D.
  Jackel, M.~Monfort, U.~Muller, J.~Zhang, et~al.
\newblock End to end learning for self-driving cars.
\newblock {\em arXiv preprint arXiv:1604.07316}, 2016.

\bibitem{chen1999experimental}
C.~Chen and H.-S. Tan.
\newblock Experimental study of dynamic look-ahead scheme for vehicle steering
  control.
\newblock In {\em Proceedings of the 1999 American Control Conference (Cat. No.
  99CH36251)}, volume~5, pages 3163--3167. IEEE, 1999.

\bibitem{chen2012kalman}
S.~Chen.
\newblock Kalman filter for robot vision: a survey.
\newblock {\em IEEE Transactions on Industrial Electronics}, 59(11):4409--4420,
  2012.

\bibitem{CosgunMCHDALTA17}
A.~Cosgun, L.~Ma, et~al.
\newblock Towards full automated drive in urban environments: {A} demonstration
  in gomentum station, california.
\newblock In {\em {IEEE} Intelligent Vehicles Symposium}, pages 1811--1818,
  2017.

\bibitem{coulter1992implementation}
R.~C. Coulter.
\newblock Implementation of the pure pursuit path tracking algorithm.
\newblock Technical report, Carnegie-Mellon UNIV Pittsburgh PA Robotics INST,
  1992.

\bibitem{Fletcher2009}
L.~Fletcher, S.~Teller, et~al.
\newblock {\em The MIT – Cornell Collision and Why It Happened}, page
  509–548.
\newblock Springer Berlin Heidelberg, 2009.

\bibitem{Fragkiadaki2016}
K.~Fragkiadaki, P.~Agrawal, et~al.
\newblock Learning visual predictive models of physics for playing billiards.
\newblock In {\em International Conference on Learning Representations (ICLR)},
  2016.

\bibitem{gal2016uncertainty}
Y.~Gal.
\newblock {\em Uncertainty in deep learning}.
\newblock PhD thesis, PhD thesis, University of Cambridge, 2016.

\bibitem{gong2004methodology}
C.~Gong and D.~McNally.
\newblock A methodology for automated trajectory prediction analysis.
\newblock In {\em AIAA Guidance, Navigation, and Control Conference and
  Exhibit}, 2004.

\bibitem{Goodfellow2016}
I.~Goodfellow, Y.~Bengio, and A.~Courville.
\newblock {\em Deep Learning}.
\newblock MIT Press, 2016.

\bibitem{resnet}
K.~He, X.~Zhang, S.~Ren, and J.~Sun.
\newblock Deep residual learning for image recognition.
\newblock In {\em Proceedings of the IEEE conference on computer vision and
  pattern recognition}, pages 770--778, 2016.

\bibitem{hochreiter1997long}
S.~Hochreiter and J.~Schmidhuber.
\newblock Long short-term memory.
\newblock {\em Neural computation}, 9(8):1735--1780, 1997.

\bibitem{huang2019uncertaintyaware}
X.~Huang, S.~McGill, B.~C. Williams, L.~Fletcher, and G.~Rosman.
\newblock Uncertainty-aware driver trajectory prediction at urban
  intersections.
\newblock 2019.

\bibitem{johnson1962folded}
N.~L. Johnson.
\newblock The folded normal distribution: Accuracy of estimation by maximum
  likelihood.
\newblock {\em Technometrics}, 4(2):249--256, 1962.

\bibitem{Kalman1960}
R.~E. Kalman.
\newblock A new approach to linear filtering and prediction problems.
\newblock {\em Transactions of the ASME--Journal of Basic Engineering},
  82(Series D):35--45, 1960.

\bibitem{kendall2017uncertainties}
A.~Kendall and Y.~Gal.
\newblock What uncertainties do we need in bayesian deep learning for computer
  vision?
\newblock In {\em Advances in Neural Information Processing Systems}, 2017.

\bibitem{kingma2014adam}
D.~P. Kingma and J.~Ba.
\newblock Adam: A method for stochastic optimization.
\newblock {\em arXiv preprint arXiv:1412.6980}, 2014.

\bibitem{Kitani2012}
K.~M. Kitani, B.~D. Ziebart, J.~A. Bagnell, and M.~Hebert.
\newblock {\em Activity Forecasting}, page 201–214.
\newblock Springer Berlin Heidelberg, 2012.

\bibitem{kong2015kinematic}
J.~Kong, M.~Pfeiffer, G.~Schildbach, and F.~Borrelli.
\newblock Kinematic and dynamic vehicle models for autonomous driving control
  design.
\newblock In {\em Intelligent Vehicles Symposium (IV), 2015 IEEE}, pages
  1094--1099. IEEE, 2015.

\bibitem{alexnet}
A.~Krizhevsky, I.~Sutskever, and G.~E. Hinton.
\newblock {ImageNet} classification with deep convolutional neural networks.
\newblock In {\em Advances in neural information processing systems}, pages
  1097--1105, 2012.

\bibitem{lakshminarayanan2017simple}
B.~Lakshminarayanan, A.~Pritzel, and C.~Blundell.
\newblock Simple and scalable predictive uncertainty estimation using deep
  ensembles.
\newblock In {\em Advances in Neural Information Processing Systems}, 2017.

\bibitem{lea2017temporal}
C.~Lea, M.~D. Flynn, R.~Vidal, A.~Reiter, and G.~D. Hager.
\newblock Temporal convolutional networks for action segmentation and
  detection.
\newblock In {\em Proceedings of the IEEE Conference on Computer Vision and
  Pattern Recognition}, pages 156--165, 2017.

\bibitem{Lee2017}
N.~Lee, W.~Choi, P.~Vernaza, C.~B. Choy, et~al.
\newblock {\em DESIRE: Distant Future Prediction in Dynamic Scenes with
  Interacting Agents}.
\newblock IEEE, Jul 2017.

\bibitem{Lefevre2014}
S.~Lefèvre, D.~Vasquez, and C.~Laugier.
\newblock A survey on motion prediction and risk assessment for intelligent
  vehicles.
\newblock {\em ROBOMECH Journal}, 1(1), Jul 2014.

\bibitem{national2017health}
NCHS.
\newblock {Health, United States, 2016: With chartbook on long-term trends in
  health}.
\newblock Technical Report 1232, National Center for Health Statistics, May
  2017.

\bibitem{ng2000algorithms}
A.~Y. Ng and S.~Russell.
\newblock Algorithms for inverse reinforcement learning.
\newblock In {\em International Conference on Machine Learning}, 2000.

\bibitem{nhtsa2017}
NHTSA.
\newblock Early estimate of motor vehicle traffic fatalities for the first half
  (jan–jun) of 2017.
\newblock Technical Report DOT HS 812 453, National Highway Traffic Safety
  Administration, December 2017.

\bibitem{ondruska2016end}
P.~Ondr{\'u}{\v{s}}ka, J.~Dequaire, D.~Z. Wang, and I.~Posner.
\newblock End-to-end tracking and semantic segmentation using recurrent neural
  networks.
\newblock {\em arXiv preprint arXiv:1604.05091}, 2016.

\bibitem{ondruvska2016deep}
P.~Ondr{\'u}{\v{s}}ka and I.~Posner.
\newblock Deep tracking: Seeing beyond seeing using recurrent neural networks.
\newblock In {\em Proceedings of the Thirtieth AAAI Conference on Artificial
  Intelligence}, pages 3361--3367. AAAI Press, 2016.

\bibitem{polzin2016implications}
S.~E. Polzin.
\newblock Implications to public transportation of emerging technologies.
\newblock 2016.

\bibitem{sandler2018inverted}
M.~Sandler, A.~Howard, M.~Zhu, A.~Zhmoginov, and L.-C. Chen.
\newblock Inverted residuals and linear bottlenecks: Mobile networks for
  classification, detection and segmentation.
\newblock {\em arXiv preprint arXiv:1801.04381}, 2018.

\bibitem{Schreier2016}
M.~Schreier, V.~Willert, and J.~Adamy.
\newblock An integrated approach to maneuver-based trajectory prediction and
  criticality assessment in arbitrary road environments.
\newblock {\em IEEE Transactions on Intelligent Transportation Systems},
  17(10):2751–2766, Oct 2016.

\bibitem{sergeev2018horovod}
A.~Sergeev and M.~D. Balso.
\newblock Horovod: fast and easy distributed deep learning in tensorflow.
\newblock {\em arXiv preprint arXiv:1802.05799}, 2018.

\bibitem{vgg}
K.~Simonyan and A.~Zisserman.
\newblock Very deep convolutional networks for large-scale image recognition.
\newblock {\em arXiv preprint arXiv:1409.1556}, 2014.

\bibitem{singh2015critical}
S.~Singh.
\newblock Critical reasons for crashes investigated in the national motor
  vehicle crash causation survey.
\newblock Technical Report DOT HS 812 115, National Highway Traffic Safety
  Administration, February 2015.

\bibitem{Smith1978}
A.~R. Smith.
\newblock Color gamut transform pairs.
\newblock In {\em Proceedings of the 5th Annual Conference on Computer Graphics
  and Interactive Techniques}, SIGGRAPH '78, pages 12--19, New York, NY, USA,
  1978. ACM.

\bibitem{Streubel2014}
T.~Streubel and K.~H. Hoffmann.
\newblock {\em Prediction of driver intended path at intersections}.
\newblock IEEE, Jun 2014.

\bibitem{wan2000unscented}
E.~A. Wan and R.~Van Der~Merwe.
\newblock The unscented kalman filter for nonlinear estimation.
\newblock In {\em Adaptive Systems for Signal Processing, Communications, and
  Control Symposium 2000. AS-SPCC. The IEEE 2000}, pages 153--158. Ieee, 2000.

\bibitem{Wang2008}
J.~Wang, D.~Fleet, and A.~Hertzmann.
\newblock Gaussian process dynamical models for human motion.
\newblock {\em IEEE Transactions on Pattern Analysis and Machine Intelligence},
  30(2):283–298, Feb 2008.

\bibitem{wang2016classifying}
K.-C. Wang and R.~Zemel.
\newblock Classifying nba offensive plays using neural networks.
\newblock In {\em Proceedings of MIT Sloan Sports Analytics Conference}, pages
  1094--1099, 2016.

\bibitem{Wiest_2017}
J.~Wiest.
\newblock {\em Statistical long-term motion prediction}.
\newblock Universität Ulm, 2017.

\bibitem{zeiler2014visualizing}
M.~D. Zeiler and R.~Fergus.
\newblock Visualizing and understanding convolutional networks.
\newblock In {\em European conference on computer vision}, pages 818--833.
  Springer, 2014.

\bibitem{Bertha2015}
J.~Ziegler, P.~Bender, M.~Schreiber, et~al.
\newblock Making bertha drive - an autonomous journey on a historic route.
\newblock {\em IEEE Intelligent Transportation Systems Magazine}, 6:8--20, 10
  2015.

\end{thebibliography}
}

\end{document}